\documentclass{article}

\usepackage{microtype}
\usepackage{graphicx}
\usepackage{subfigure}
\usepackage{booktabs} 

\usepackage{hyperref}

\usepackage[accepted]{icml2018}

\usepackage{bbm}
\usepackage{amsthm}
\usepackage{amsmath}
\usepackage{amssymb}
\usepackage{color}
\usepackage{xspace}
\usepackage{pgfplots, pgfplotstable}

\def\deriv{\mathrm{d}}
\def\grad{\nabla}

\newcommand{\cov}[2]{\lVert#1\rVert_{#2}^2}

\renewcommand{\mid}[0]{\,\vert\,}
\newcommand{\at}[2][]{#1|_{#2}}

\def\ouralg{Smoothie\xspace}

\def\meanparam{\theta}
\def\varparam{\phi}
\def\qparam{w}

\def\expected{\mathbb{E}}

\def\dataset{\mathcal{D}}
\def\S{\mathcal{S}}
\def\A{\mathcal{A}}
\def\N{\mathcal{N}}

\def\calL{\mathcal{L}}

\def\Real{\mathbb{R}}
\def\mean{\mu}
\def\var{\Sigma}

\def\qexp{Q^\pi}
\def\qdet{Q^\pi}
\def\qtilde{\tilde{Q}^\pi}
\def\atilde{\tilde{a}}
\def\rtilde{\tilde{r}}
\def\stilde{\tilde{s}}

\def\eps{\epsilon}

\def\eg{{\em e.g.,\xspace}}
\def\ie{{\em i.e.,\xspace}}

\def\wrt{{\em w.r.t.}}

\newcommand{\figref}[1]{Figure~\ref{#1}}

\newcommand{\tabref}[1]{Table~\ref{#1}}

\newcommand{\secref}[1]{Section~\ref{#1}}

\def\objmr{O_{\text{ER}}}
\def\objdpg{J_{DPG}}
\def\objtilde{O_{\text{ER}}}
\def\objtrust{O_{\text{TR}}}

\newcommand{\trans}[1]{{#1}^{\ensuremath{\mathsf{T}}}}
\newcommand{\kl}[2]{{\mathrm{KL}}\left(#1~\Vert~#2\right)}

\icmltitlerunning{Smoothed Action Value Functions for Learning Gaussian Policies}

\begin{document}

\twocolumn[
\icmltitle{
  Smoothed Action Value Functions
  for Learning Gaussian Policies
}



\icmlsetsymbol{equal}{*}

\begin{icmlauthorlist}
\icmlauthor{Ofir Nachum}{goo}
\icmlauthor{Mohammad Norouzi}{goo}
\icmlauthor{George Tucker}{goo}
\icmlauthor{Dale Schuurmans}{goo,alb}
\end{icmlauthorlist}

\icmlaffiliation{goo}{Google Brain}
\icmlaffiliation{alb}{Department of Computing Science, University of Alberta}

\icmlcorrespondingauthor{Ofir Nachum}{ofirnachum@google.com}

\icmlkeywords{Machine Learning, ICML, Reinforcement Learning, Artificial Intelligence}

\vskip 0.3in
]



\printAffiliationsAndNotice{}  

\begin{abstract}
State-action value functions (\ie\ Q-values) are ubiquitous
in reinforcement learning (RL), giving rise to popular
algorithms such as SARSA and Q-learning.
We propose a new notion of action value defined by a Gaussian
smoothed version of the expected Q-value.
We show that such smoothed Q-values still satisfy a Bellman equation,
making them learnable from experience 
sampled from an environment.
Moreover, the gradients of expected reward with respect to the
mean and covariance of a parameterized Gaussian policy can be 
recovered from the gradient and Hessian of the smoothed Q-value function.
Based on these relationships, we develop new algorithms for training
a Gaussian policy directly from a learned smoothed Q-value approximator.
The approach is additionally amenable to proximal optimization by
augmenting the objective with a penalty on KL-divergence from a previous policy.
We find that the ability to learn both a mean and covariance during training
leads to significantly improved
results on standard continuous control benchmarks.
\end{abstract}

\comment{
Algorithms which make use of gradients of Q-value functions
to learn an agent's behavioral policy are popular for
their ability to avoid the high variance in updates based
on Monte Carlo rollouts used in standard policy gradient methods.
However, these algorithms often heavily restrict the expressivity of the policy,
confining it to be either deterministic or Gaussian with an un-trainable covariance.
Thus, these algorithms exhibit poor exploration and are not easily amenable
to trust region or proximal policy methods.
We present a new notion of Q-values that can be interpreted as 
a Gaussian-smoothed version of the standard expected Q-values.
We show that gradients of expected reward with respect to the
mean and covariance of a parameterized Gaussian policy may be 
expressed in terms of the gradient and Hessian of the Q-values
with respect to mean action.
Thus, we derive an algorithm of training both mean and covariance
using a learned Q-value function.
Our algorithm is easily amenable to proximal policy techniques by
augmenting the objective with a penalty on KL-divergence from a previous policy.
We show that the ability of our algorithm to learn both mean and covariance
during training leads to good results on standard continuous control benchmarks.
}

\section{Introduction}

Model-free reinforcement learning algorithms often alternate between
two concurrent but interacting processes: 
(1) {\em policy evaluation},
where an {\em action value function} (\ie\ a Q-value)
is updated to obtain 
a better estimate of the return associated with taking a specific action, 
and (2) {\em policy improvement},
where the policy is updated aiming to maximize the current value function. 
In the past, different notions of Q-value have led to distinct but
important families of RL methods.
For example,
SARSA~\citep{sarsa1, suttonbook, sarsa2}
uses the {\em expected} Q-value, defined as the expected
return of following the current policy.
Q-learning~\citep{qlearning} exploits a {\em hard-max} notion of Q-value,
defined as the expected return of following an optimal policy.
Soft Q-learning~\citep{haarnoja2017reinforcement} and PCL~\citep{pcl}
both use a {\em soft-max} form of Q-value, defined as the future return of 
following an optimal entropy regularized policy.
Clearly,
the choice of Q-value function has a considerable effect on the resulting
algorithm;
for example,
restricting the types of policies that can be expressed,
and
determining the type of exploration that can be naturally applied.
In each case, the Q-value at a state $s$ and action $a$ answers the question, 

{\em ``What would my future value from $s$ be if I were to take
an initial action $a$?"}  

Such information about a hypothetical action is helpful when learning a policy;
we want to nudge the policy distribution to favor actions with potentially
higher Q-values.

In this work, we investigate the practicality and benefits of answering
a more difficult, but more relevant, question: 

{\em ``What would my future value from $s$ be if I were to 
sample my initial action from a distribution 
centered at $a$?"} 

We focus our efforts on Gaussian policies and 
thus the 
counterfactual posited by the Q-value 
inquires about the expected future return of following the policy
when changing
the mean of the initial Gaussian distribution.
Thus, our new notion of Q-values 
maps a state-action pair $(s, a)$ 
to the expected return of first taking an action sampled from
a normal distribution $N(\cdot|a, \var(s))$ centered at $a$,
and following actions sampled from the current policy thereafter. 
In this way,
the Q-values we introduce may be  interpreted as a
Gaussian-smoothed version of the expected Q-value,
hence we term them {\em smoothed} Q-values.

We show that smoothed Q-values possess a number of important
properties that make them attractive for use in RL algorithms.
It is clear from the definition of smoothed Q-values that,
if known, their structure is highly beneficial for learning
the mean of a Gaussian policy.
We are able to show more than this:
although the smoothed Q-values are not a direct function
of the covariance, one can surprisingly use knowledge
of the smoothed Q-values to derive updates to
the covariance of a Gaussian policy.
Specifically, the gradient of the standard expected return objective
with respect to the mean and covariance of a Gaussian policy is equivalent
to the gradient and Hessian of the smoothed Q-value function, respectively.
Moreover, we show that the smoothed Q-values 
satisfy a single-step Bellman consistency,
which allows bootstrapping to be used to train them via function approximation.

These results lead us to propose an algorithm, {\em \ouralg},
which,
in the spirit of (Deep) Deterministic Policy Gradient (DDPG)~\citep{ddpg1, ddpg2},
trains a policy using the derivatives of a trained (smoothed) Q-value function
to learn a Gaussian policy.
Crucially, unlike DDPG, which is restricted to deterministic policies
and is well-known to have
poor exploratory behavior~\citep{haarnoja2017reinforcement},
the approach we develop
is able to utilize a non-deterministic
Gaussian policy parameterized by both a mean and a covariance,
thus allowing the policy to be exploratory by default
and alleviating the need for excessive hyperparameter tuning.
On the other hand,
compared to standard policy gradient algorithms~\citep{williams1991function,konda2000actor},
Smoothie's utilization of the derivatives of a Q-value function
to train a policy
avoids the high variance and sample inefficiency of stochastic updates. 

Furthermore, we show that \ouralg can be easily adapted to incorporate
proximal policy optimization techniques by augmenting the objective
with a penalty on KL-divergence from a previous version of the policy.
The inclusion of a KL-penalty is not feasible in the standard DDPG
algorithm, but we show that it is possible with our formulation, and it
significantly improves stability and overall performance.
On standard continuous control benchmarks, our results are competitive with or exceed
state-of-the-art, especially for more difficult tasks
in the low-data regime.

\comment{
For most RL algorithms including SARSA, expected SARSA, and
Q-leanring, the policy improvement step is straightforward: An
improved policy $\pi$ is obtained by taking the greedy action $a$ that
maximizes the action value function $Q(s, a)$, or alternatively, the
greedy action is taken with probability $1-\epsilon$ and a random
action is taken with probability $\epsilon$. Using such
$\epsilon$-greedy schemes may improve exploration and facilitate
formulating convergence guarantees. However, policy improvement by
finding the locally optimal policy for an arbitrary action value
function is not tractable, especially in continuous and
high-dimensional action spaces. For example, if one considers the
family of multivariate guassian policies in a continuous action space,
then policy optimization within the feasible policies is not
straighforward.

Prior work on deterministic policy gradient (DPG) and its extensions
substitude the mean of a gaussian policy into the action value
function and make use of the gradient of the action value function to
update the mean of a gaussian policy to improve action values
estimates. Such formulations are interesting because they can handle
off-policy data, and they do not require Monte Carlo sampling to
compute expected Q-values under the current policy, resulting in a
learning algorithm with a lower variance. However, DPG and its
variants make a strong assumption about the policy. They assume that
the policy is deterministic in the limit that the variance of
Gaussians goes to zero. It was thought that the one needs to assume
policy determinism to achieve the benefits of the DPG formulation
(???), but in this paper we show the policy determinism assumption is
not necessary. In other words, one can keep the benefits of DPG in
making use of action value gradients and off-policy data, but also
allow for using a general family of stochastic Gaussian policies.

Model-free reinforcement learning (RL) aims to optimize an
agent's behavior policy through trial and error interaction with a
black box environment.  An agent alternates between
observing a \emph{state} provided by the environment 
(\eg\ joint positions and velocities),
applying an \emph{action} to the environment 
(\eg\ force or torque),
and receiving a \emph{reward} from the environment
(\eg\ velocity in a specific desired direction).
The agent's objective is to maximize
the long term sum of rewards received from the environment.

Within continuous control, the agent's policy is traditionally 
parameterized by a uni-model Gaussian.  The mean and covariance
defining the Gaussian are then trained using policy gradient
methods~\citep{konda2000actor,williams1991function}
to maximize expected total reward.  Policy gradient uses experience
sampled stochastically from its policy
as a cue to nudge the mean and covariance
to put more probability mass on experience that yielded 
a higher long term reward
than expected and vice versa for experience
that yielded a lower long term reward.  
The stochastic nature of this training paradigm makes policy gradient
methods unstable and exhibit high variance.  
To mitigate this problem, large batch sizes or 
trust region methods~\citep{trpo,tpcl,ppo}
are utilized, although both of these remedies can require
collecting a large amount of experience, making them infeasible
for real-world applications.

As an attempt to circumvent the issue of high variance due to 
highly stochastic updates, recent years have introduced policy
gradient algorithms which update policy parameters based only
on the surface of a learned Q-value function.  
The most widely used such algorithm is (Deep) Deterministic
Policy Gradient (DDPG)~\citep{ddpg1, ddpg2}.
In DDPG, the policy is deterministic, parameterized only by a mean.
A Q-value function is trained to take in a state and action 
and return the future discounted sum of rewards of first taking
the specified action and subsequently following the deterministic 
policy.  Thus, the suface of the learned Q-value function 
dictates how the policy should be updated - along
the gradient of the Q-value with respect to the input action.

While DDPG has successfully avoided the highly-stochastic policy
updates associated with traditional policy gradient algorithms,
its deterministic policy naturally leads to poor exploration.
A deterministic policy gives no indication regarding which directions
in action space to explore.  Thus, in practice, the trained policy
differs from the behavior policy, which is augmented with Gaussian
noise whose variance is treated as a hyperparameter to optimize.  
Even so, DDPG is well-known to have poor exploratory behavior~\citep{haarnoja2017reinforcement}.

In this paper, we present a method which applies the same technique
of updating a policy based only on a learned Q-value function
to a \emph{stochastic} policy parameterized by both a mean and 
covariance.  
We show that given access to the proposed Q-value function or 
a sufficiently accurate approximation, it is possible to derive
unbiased updates to both the policy mean and covariance.
Unlike recent attempts at this (\eg~\citet{epg}), our updates
require neither approximate integrals nor low-order assumptions on
the form of the true Q-value.
Crucially, providing a method to update the covariance
allows the policy to be exploratory by default and alleviates the 
need for excessive hyperparameter tuning.  Moreover, we
show that our technique can be easily adapted to incorporate
trust region methods by 
augmenting the objective with a penalty
on KL-divergence from a previous version of the policy.
The inclusion of a trust region is not possible in standard
DDPG and we show that it
improves stability and overall performance 
significantly when our
algorithm is evaluated on standard continuous control benchmarks.
}

\section{Formulation}
\label{prelim}

We consider the standard model-free RL problem represented a Markov
decision process (MDP), consisting of a state space $\S$ and an
action space $\A$.  At iteration $t$ the agent encounters a state
$s_t\in\S$ and emits an action $a_t\in\A$, after which the environment
returns a scalar reward $r_t\sim R(s_t, a_t)$ and places the agent in
a new state $s_{t+1}\sim P(s_t, a_t)$. 

We focus on continuous control tasks, where the actions are
real-valued, \ie~$\A \equiv \Real^{d_a}$. Our observations at a state
$s$ are denoted $\Phi(s) \in \Real^{d_s}$. We parameterize the
behavior of the agent using a stochastic policy $\pi(a \mid s)$, which
takes the form of a Gaussian density at each state $s$.
The Gaussian policy is parameterized by a mean and a covariance function,
$\mean(s): \Real^{d_s} \to \Real^{d_a}$ and
$\var(s): \Real^{d_s} \to\Real^{d_a}\times\Real^{d_a}$
so that $\pi(a \mid s)= \N(a \mid \mean(s),\var(s))$,
where
\begin{equation}
N(a \mid \mean,\var) ~=~ |2\pi\var|^{-1/2} \exp\left\{-\frac{1}{2} \cov{a - \mean}{\var^{-1}}\right\},
\end{equation}
here using the notation $\cov{v}{A} = \trans{v} A v$. 

\comment{
Below we develop new RL training methods for
this family of parametric policies, but some of the ideas presented
may generalize to other families of policies as well. 
We begin the
formulation by reviewing some prior work on learning Gaussian policies.
}

\subsection{Policy Gradient for Generic Stochastic Policies}

The optimization objective (expected discounted return), as a function
of a generic stochastic policy, is expressed in terms of the expected
action value function $\qexp(s, a)$ by,
\begin{equation}
  \objmr(\pi) ~=~
  \int_\S \int_\A \pi(a \mid s) \qexp(s, a) \,\deriv a\, \deriv \rho^{\pi}(s)~,
\label{eq:objmr}
\end{equation}
where $\rho^\pi(s)$ is the state visitation distribution under
$\pi$, and $\qexp(s, a)$ is recursively defined using the Bellman equation,
\begin{equation}
\qexp(s, a) = \expected_{r,s'}\left[
r + \gamma\! \int_{\A} \qexp(s',a') \pi(a' \mid s') \,\deriv a \right]~,
\label{eq:qexp-bell}
\end{equation}
where $\gamma \in [0, 1)$ is the discount factor.
For brevity, we suppress explicit denotation of the
distribution $R$ over immediate rewards and $P$ over state transitions.

The policy gradient theorem~\citep{sutton2000policy} expresses the
gradient of $\objmr(\pi_{\theta})$ \wrt~$\theta$, the tunable
parameters of a policy $\pi_{\theta}$, as,
\begin{eqnarray}
\lefteqn{
\grad_\theta \objmr(\pi_\theta) =
\int_\S \int_\A
\grad_\theta \pi_{\theta}(a\mid s)\qexp(s, a)\, \deriv a\,\deriv\rho^\pi(s)~
}
\nonumber
\\
\! & \!=\! & \!
\int_\S \, \expected_{a \sim \pi_\theta(a\!\mid\!s)} \left[
\grad_\theta \log \pi_{\theta}(a \mid s) \qexp(s, a) \right]\deriv \rho^\pi(s)
.
\quad
~~
\label{eq:pg-thrm}
\end{eqnarray}
In order to approximate the expectation on the RHS of
\eqref{eq:pg-thrm}, one often resorts to an empirical average over
on-policy samples from $\pi_{\theta}(a \mid s)$. This sampling scheme
results in a gradient estimate with high variance, especially when
$\pi_{\theta}(a \mid s)$ is not concentrated. Many policy gradient
algorithms, including actor-critic variants, trade off variance and
bias, \eg~by attempting to estimate $\qexp(s, a)$ accurately using
function approximation and the Bellman equation. 

\comment{
  In the simplest
  scenario, an unbiased estimate of $\qexp(s, a)$ is formed by
  accumulating discounted rewards from each state forward using a single
  Monte Carlo sample.
}

\comment{
The objective of the policy is to maximize expected future discounted reward
at each state until reaching some terminal time $T\in\mathbb{N}\cup\{0,\infty\}$:
\begin{equation}
\objmr(s_0) = \expected_{a_i\sim N(\mean(s_i),\var(s_i)),r_i,s_{i+1}}
\left[\sum_{i=0}^T \gamma^i r_i
\right].
\end{equation}
The objective may also be expressed in terms of \emph{expected} Q-values as
\begin{equation}
\objmr(s_0,\mean,\var) = \int_\A \qexp(s_0, a) N(a|\mean(s_0),\var(s_0)) \deriv a,
\end{equation}
where $\qexp$ is defined recursively as
\begin{equation}
\qexp(s,a) = \expected_{r,s'}\left[
r + \gamma \int_{\A} \qexp(s',a') N(a'|\mean(s'),\var(s')) \deriv a' \right],
\end{equation}
to represent the expected future discounted reward of taking action $a$ at state $s$
and subsequently following the policy $\mean,\var$.

The state-agnostic objective is then
\begin{equation}
\objmr(\mean,\var) = \int_\S \int_\A \qexp(s,a) N(a|\mean(s),\var(s)) \deriv a \deriv \rho(s),
\label{eq:objmr}
\end{equation}
where $\rho$ is the state distribution induced by the policy $\mean,\var$.
}

\subsection{Deterministic Policy Gradient}

\citet{ddpg1} study the policy gradient for the specific class of
Gaussian policies in the limit where the policy's covariance
approaches zero. 
In this scenario, the policy becomes deterministic
and samples from the policy approach the Gaussian mean. Under a
deterministic policy $\pi\equiv (\mu,\var\to0)$, one can estimate the
expected future return from a state $s$ as,
\begin{equation}
\lim_{\var\to 0} \int_\A \pi(a \mid s) \qexp(s, a)\,\deriv a ~=~
\qexp(s, \mu(s))~.
\end{equation}
Accordingly, \citet{ddpg1} express the gradient of the expected
discounted return objective for $\pi_\theta\equiv \delta(\mean_\theta)$ as,
\begin{equation}
  \grad_\theta \objmr(\pi_\theta)
\!=\!
  \int_\S 
  \frac{\partial \qexp(s,a)}{\partial a}\at[\Big]{a=\mean_\theta(s)}
\!\!
  \grad_\theta \mean_\theta(s) \deriv \rho^\pi(s)
.
  \label{eq:dpg-thrm}
\end{equation}
This characterization of the policy gradient theorem for
deterministic policies is called {\em deterministic policy gradient} (DPG).
Since no Monte Carlo sampling is required for estimating the gradient,
the variance of the estimate is reduced.
On the other hand, the deterministic nature of the policy
can lead to poor exploration and training instability in practice.

In the limit of $\var \to 0$, one can also re-express the Bellman
equation~\eqref{eq:qexp-bell} as,
\begin{equation}
\qexp(s,a) ~=~ \expected_{r,s'} \left[r + \qexp(s', \mean(s')) \right]~.
\end{equation}
Therefore, a value function approximator $\qexp_{\qparam}$ can be optimized by
minimizing the Bellman error,
\begin{equation}
  E(w) = \sum_{(s,a,r,s') \in \dataset}(\qexp_\qparam(s,a) - r - \gamma \qexp_\qparam(s', \mean(s'))^2~,
  \label{eq:dq-bellman}
\end{equation}
for transitions $(s,a,r,s')$ sampled from a dataset $\dataset$ of
interactions of the agent with the environment.
The deep variant of DPG known as DDPG~\citep{ddpg2} alternates between
improving 
the action
value estimate
by gradient descent on \eqref{eq:dq-bellman} and improving the policy
based on~\eqref{eq:dpg-thrm}. 

To improve sample efficiency, \citet{degris2012} and \citet{ddpg1}
replace the state visitation distribution $\rho^\pi(s)$
in \eqref{eq:dpg-thrm} with an off-policy visitation distribution
$\rho^\beta(s)$ based on a {\em replay buffer}. This subsititution
introduces some bias in the gradient estimate 
\eqref{eq:dpg-thrm}, but
previous work has found that it works well in practice and
improves the sample efficiency of the policy gradient algorithms.
We also adopt a similar heuristic in our method to make use of off-policy data.

In practice, DDPG exhibits improved sample efficiency over
standard policy gradient algorithms:
using off-policy data to train Q-values
while basing policy updates on their gradients
significantly improves
stochastic
policy updates dictated by~\eqref{eq:pg-thrm},
which require a large number of samples to reduce noise.
On the other hand, the deterministic nature of the policy learned by DDPG
leads to poor exploration and instability in training.
In this paper, we propose an algorithm which, like DDPG,
utilizes derivative information of learned Q-values for better
sample-efficiency, 
but which, unlike DDPG, is able to learn a Gaussian policy
and imposes a KL-penalty for better exploration and stability.

\comment{
The objective for $\mean$ becomes
\begin{equation}
\objdpg(\mean) = \lim_{\var\to0} \objmr(\mean,\var) = \int_\S \qdet(s,\mean(s)) \deriv \rho(s).
\end{equation}
A parameterized $\mean_\phi$ can thus be trained according to 
\begin{equation}
\Delta\phi \propto \int_\S \frac{\partial \qdet(s,a)}{\partial a}\at[\big]{a=\mean_\phi(s)} \nabla_\phi \mean_\phi(s) \deriv \rho(s).
\end{equation}
}

\comment{
\subsection{Stochastic Value Gradients for Gaussian Policies}

Inpired by deterministic policy gradients, \citet{heess2015learning}
propose to reparameterize the expectation \wrt~a Guassian policy in
\eqref{eq:pg-thrm} with an expectation over a noise variable
$\epsilon$ drawn from a standard normal distribution. Note that a
stochastic action drawn from a Gaussian policy $a \sim
\N(\mean(s),\var(s))$ can be reparameterized as $a = \mean(s) +
L(s)\eps$, for $\eps \sim \N(0, I_{d_a})$ and $L(s) \trans{L(s)} =
\var(s)$. Accordingly, the policy gradients take the form of,
\begin{equation}
\begin{aligned}
  \grad&_\theta \objmr(\pi_\theta) =\\
  &\int_\S \, \expected_{\epsilon}
  \frac{\partial \qexp(s,a)}{\partial a}\at[\big]{a=\mean_\theta + L_{\theta}\epsilon}
  [\grad_\theta \mean_\theta + \grad_\theta L_\theta\epsilon] \deriv \rho^\pi(s)~,
\end{aligned}
\label{eq:pg-svg}
\end{equation}
where for brevity, we dropped the dependence of $\mean_\theta(s)$ and
$L_\theta(s)$ on $s$. Similarly, one can re-express Bellman equations
using an expectation over the noise variable.

The key advantage of this formulation by \citet{heess2015learning},
called stochastic value gradients (SVG), over generic policy gradients
\eqref{eq:pg-thrm} is that similar to DPG~\eqref{eq:dpg-thrm}, SVG
makes direct use of the gradient of the Gaussian mean functions with
respect to the model parameters. The benefit over DPG is that SVG
keeps the policy stochastic and enables learning the covariance
function at every state, but the key disadvantage over DPG is that SVG
requires sampling from a noise variable to estimate the gradients,
resulting in a higher variance.

In this paper, we show how one can combine the benefits of DPG and SVG
to formulate policy gradients for Gaussian policies wihtout requiring
to sample from a noise variable to estimate the gradients, hence a
lower variance.
}

\begin{figure}[t]
\begin{center}
  \begin{tabular}{c}
    \includegraphics[width=0.67\columnwidth]{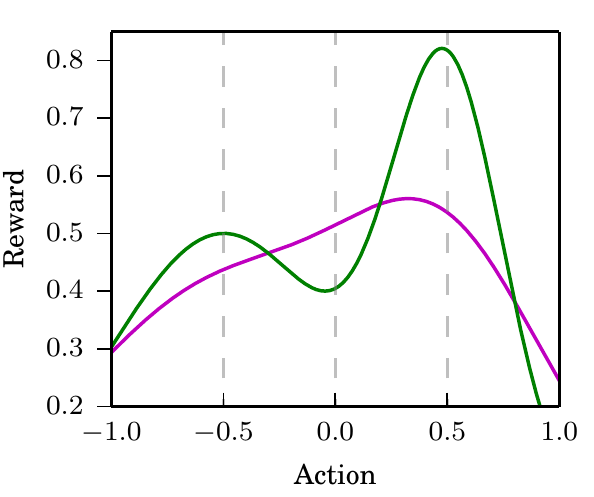} \\
    \multicolumn{1}{c}{\vspace{-0.2in}~\includegraphics[width=0.8\columnwidth]{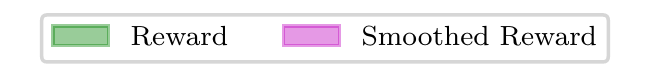}}
  \end{tabular}
\end{center}
\caption{
A simple expected reward function, shown in green,
with a Gaussian-smoothed version, shown in magenta.
}
\vspace{-0.2in}
\label{fig:bumps}
\end{figure}

\section{Idea}
\label{sec:idea}

Before giving a full exposition,
we use a simplified scenario to illustrate 
the key 
intuitions behind the
proposed approach and how it differs fundamentally from previous methods.

Consider a one-shot decision making problem over a one dimensional
action space with a single state.
Here the expected reward is given by a function over the real line,
which also corresponds to the optimal Q-value function;
\figref{fig:bumps} gives a concrete example.
We assume the policy $\pi$ is specified by a Gaussian distribution 
parameterized by a scalar mean $\mu$ and standard deviation $\sigma$.
The goal is to optimize the policy parameters to maximize expected reward.

A naive policy gradient method updates the parameters by sampling
$a\sim\pi$, observing reward $r_i$,
then adjusting $\mu$ and $\sigma$
in directions
$\Delta\mu=\frac{d\log\pi(a_i)}{d\mu}r_i=\frac{\mu-a_i}{\sigma^2}r_i$
and
$\Delta\sigma=\frac{d\log\pi(a_i)}{d\sigma}r_i
=\big(\frac{(\mu-a_i)^2}{\sigma^2}-\frac{1}{\sigma}\big)r_i$.
Note that
such updates suffer from large variance, particularly when $\sigma$ is small.

To reduce the variance of direct policy gradient,
deterministic policy gradient methods leverage a value function approximator $\qexp_w$,
parameterized by $w$, to approximate $\qexp$.
For example, in this scenario, 
vanilla DPG would sample an action $a_i=\mu+\varepsilon_i$
with exploration noise $\varepsilon_i\sim N(0,\sigma^2)$,
then update $\mu$ using
$\Delta\mu=\frac{\partial \qexp_w(a)}{\partial a}\at[\big]{a=\mu}$
and $\qexp_w$ using $\Delta w=(r_i - \qexp_w(a_i))\nabla_w \qexp_w(a_i)$.
Clearly, this update exhibits reduced variance,
but requires $Q^\pi_w$ to approximate $Q^\pi$
(the green curve in \figref{fig:bumps})
to control bias.
Unfortunately, DPG is not able to learn the exploration variance $\sigma^2$.
Variants of DPG such as SVG~\cite{heess2015learning} and EPG~\cite{epg}
have been proposed to work with stochastic policies.
However they 
either 
have restrictive assumptions on the form of the true Q-value,
introduce a noise into the policy updates, 
or require an approximate integral,
thus losing the advantage of deterministic gradient updates.
\comment{
By contrast, SVG is able to work with a stochastic policy and learn
the variance, but loses the advantage of deterministic gradient updates.
In particular, vanilla SVG would sample $\varepsilon_i\sim N(0,1)$ 
then update $\mu$ using $\Delta\mu=\frac{\qexp_w(\mu+\sigma\varepsilon_i)}{da}$,
$\sigma$ using 
$\Delta\sigma=\frac{\qexp_w(\mu+\sigma\varepsilon_i)}{da}\varepsilon_i$,
and $\qexp_w$ using 
$\Delta w=\frac{d\qexp_w(\mu+\sigma\varepsilon_i)}{dw}
(\qexp_w(\mu+\sigma\varepsilon_i)-r_i)$,
all of which reintroduce significant variance in the updates.
}

Note, however, that the expected value at any given location
is actually given by a \emph{convolution} of the Gaussian policy
with the underlying expected reward function.
Such a process inherently smooths the landscape,
as shown in the magenta curve in \figref{fig:bumps}.
Unfortunately, DPG completely ignores this smoothing effect 
by trying to approximate $Q^\pi$,
while policy gradient methods 
only benefit from it indirectly through sampling.
A key insight is that this smoothing effect can be captured directly
in the value function approximator itself,
bypassing any need for sampling 
or approximating $Q^\pi$.
That is, instead of using an approximator to model $\qexp$,
one can directly approximate the smoothed version given by
$\qtilde(a) = \int_{\A}N(\atilde|a,\sigma^2)\qexp(\atilde)\:d\atilde$
(the magenta curve in \figref{fig:bumps}),
which, crucially, satisfies $\objmr(\pi) = \qtilde(\mu)$.

Based on this observation, we propose a novel actor-critic strategy 
below that uses a function approximator $\qtilde_w$ to model $\qtilde$.
Although approximating $\tilde Q^\pi$ instead of $Q^\pi$ might appear
to be a subtle change,
it is a major alteration to existing actor-critic approaches.
Not only is approximating the magenta curve in \figref{fig:bumps}
far easier than the green curve,
modeling $\tilde Q^\pi$ allows the policy parameters to be updated 
\emph{deterministically} for any given action.
In particular,
in the simple scenario above,
if one sampled an action from the current policy, $a_i\sim\pi$,
and observed $r_i$,
then $\mu$ could be updated using 
$\Delta\mu=\frac{\partial \qtilde_w(a)}{\partial a}\at[\big]{a=\mu}$,
$\sigma$ using $\Delta\sigma=\frac{\partial^2\qtilde_w(a)}{\partial a^2}\at[\big]{a=\mu}$
(a key result we establish below),
and $\qtilde_w$ using
$\Delta w=(r_i - \qtilde_w(\mu))\nabla_w \qtilde_w(\mu)$.

Such a strategy combines the best aspects of DPG and policy gradient 
while conferring additional advantages:
(1) the smoothed value function $\qtilde$ cannot add but can only remove local 
minima from $\qexp$;
(2) $\qtilde$ is smoother than $\qexp$ hence easier to approximate;
(3) approximating $\qtilde$ allows deterministic gradient updates for $\pi$;
(4) approximating $\qtilde$ allows gradients to be computed for both the mean and variance parameters.
Among these advantages, 
DPG shares only 3 and
policy gradient only 1.
We will see below that the new strategy we propose
significantly outperforms existing approaches,
not only in the toy scenario depicted in \figref{fig:bumps},
but also in challenging benchmark problems.

\section{Smoothed Action Value Functions}
\label{sec:method}


Moving beyond a simple illustrative scenario,
the key contribution of this paper is to introduce the general notion of a
{\em smoothed action value function},
the gradients of which provide an effective signal for optimizing the
parameters of a Gaussian policy.
Smoothed Q-values, which we denote $\qtilde(s, a)$,
differ from ordinary Q-values $\qexp(s, a)$ 
by not assuming the first action of the agent is fully specified;
instead, they assume only that a Gaussian centered at the action is known.
Thus, to compute $\qtilde(s, a)$, one has to
perform an expectation of $\qexp(s, \atilde)$ for actions $\atilde$
drawn in the vicinity of $a$. More formally, smoothed action values
are defined as,
\begin{equation}
\qtilde(s,a) ~=~ \int_{\A} N(\atilde \mid a,\var(s))\,\qexp(s, \atilde)
\:\deriv\atilde~.
\label{eq:qtilde1}
\end{equation}
With this definition of $\qtilde$, one can re-express the gradient of
the expected
reward objective (Equation~\eqref{eq:pg-thrm}) for a Gaussian policy $\pi\equiv(\mean,\var)$ as,
\begin{equation}
\nabla_{\mu,\Sigma} ~\objmr(\pi) = \int_\S \nabla_{\mu,\Sigma} ~\qtilde(s, \mean(s))\:\deriv\rho^\pi(s)~.
\label{eq:objtilde}
\end{equation}
The insight that differentiates this approach from prior work 
\citep{heess2015learning, epg}
is that instead of learning a function
approximator for $\qexp$ then drawing samples to
approximate the expectation in \eqref{eq:qtilde1} and its derivative,
we directly learn a function approximator for $\qtilde$.

One of the key observations
that enables learning a function approximator for $\qtilde$
is that smoothed Q-values satisfy a notion of Bellman consistency.
First, note that for Gaussian policies $\pi\equiv(\mean,\var)$ we have
the following relation between the expected and smoothed Q-values:
\begin{equation}
\qexp(s, a) = \expected_{r, s'}[r + \gamma \qtilde(s', \mean(s'))]~.
\label{eq:qexp-tilde}
\end{equation}
Then, combining \eqref{eq:qtilde1} and~\eqref{eq:qexp-tilde}, one can derive
the following one-step Bellman equation for smoothed Q-values,
\begin{equation}
\qtilde(s, a) \!=\!\! \int_{\A} \!\!
N(\atilde \mid a,\var(s))\,\expected_{\rtilde,\stilde'}
\!
\left[\rtilde + \gamma \qtilde(\stilde', \mean(\stilde')) \right]
\deriv\atilde,
\label{eq:qtilde2}
\end{equation}
where $\rtilde$ and $\stilde'$ are sampled from $R(s,\atilde)$ and
$P(s,\atilde)$. 
Below, we elaborate on how one can make use of
the derivatives of $\qtilde$ to learn $\mu$ and $\var$, and how the
Bellman equation in \eqref{eq:qtilde2} enables direct optimization
of $\qtilde$.

\subsection{Policy Improvement} 

We assume
a Gaussian policy
$\pi_{\meanparam,\varparam} \equiv (\mean_{\meanparam},\var_{\varparam})$
parameterized by
$\meanparam$ and $\varparam$ for the mean and the covariance respectively.
The gradient of the objective \wrt~the mean parameters
follows from the policy gradient theorem in conjunction with
\eqref{eq:objtilde} 
and is almost identical to \eqref{eq:dpg-thrm},
\begin{equation}
\nabla_{\meanparam} \objmr(\pi_{\meanparam,\varparam}) 
~=~ 
\int_\S
\frac{\partial \qtilde(s, a)}{\partial a}\at[\Big]{a=\mean_{\meanparam}(s)} 
\nabla_{\meanparam} \mean_{\meanparam}(s) \deriv \rho^\pi(s).
\label{eq:mean-upd}
\end{equation}

Estimating the derivative of the objective \wrt~the covariance parameters
is not as straightforward, since $\qtilde$ is not a direct function of $\var$.
However, a key result
is that the second derivative of
$\qtilde$ \wrt~actions is sufficient to exactly compute the derivative of
$\qtilde$ \wrt~$\var$.
\begin{eqnarray}
\textbf{Theorem 1.}
\quad
\displaystyle
\frac{\partial \qtilde(s, a)}{\partial \var(s)}
=
\frac{1}{2} \cdot \frac{\partial^2 \qtilde(s, a)}{\partial a^2}
\;\;\forall s, a
~.
~~~
\label{eq:sigma-trick}
\end{eqnarray}
A proof of this identity is provided in the Appendix. 
%
The full derivative \wrt~$\varparam$ 
can then be shown to 
take the form,
\begin{equation}
\nabla_{\varparam} \objmr(\pi_{\meanparam,\varparam}) 
=
\frac{1}{2}
\!
\int_\S
\!
\frac{\partial^2 \qtilde(s, a)}{\partial a^2}\at[\Big]{a=\mu_{\meanparam}(s)} 
\!\!
\nabla_{\varparam} \var_{\varparam}(s) \deriv \rho^\pi(s).
\label{eq:var-upd}
\end{equation}

\subsection{Policy Evaluation}
\label{sec:train-q}

There are two natural ways to optimize $\qtilde_\qparam$.  
The first approach leverages~\eqref{eq:qtilde1}
to update $\qtilde$ based on the expectation of $\qexp$.
In this case, one first trains a parameterized model $\qexp_{\qparam}$
to approximate the standard $\qexp$ function using conventional methods
\citep{sarsa1,  suttonbook, sarsa2},
then fits $\qtilde_\qparam$ to $\qexp_\qparam$ based on \eqref{eq:qtilde1}.
In particular, 
given transitions $(s, a, r, s')$ sampled from interactions with the
environment, one can train $\qexp_\qparam$ to minimize the Bellman error
\begin{equation}
(\qexp_\qparam(s, a) - r - \gamma \qexp_\qparam(s', a'))^2,
\end{equation}
where $a'\sim N(\mean(s'),\var(s'))$.
Then, $\qtilde_\qparam$ can be optimized to minimize the squared error
\begin{equation}
(\qtilde_\qparam(s, a) - \expected_{\atilde}[\qexp_\qparam(s, \atilde)])^2,
\end{equation}
where $\atilde\sim N(a, \var(s))$, using several samples.
When the target values in the residuals
are treated as fixed (\ie~using a target network),
these updates will reach a fixed point when $\qtilde_\qparam$ satisfies
the recursion in the Bellman equation \eqref{eq:qtilde1}.

The second approach requires a single function approximator for
$\qtilde_\qparam$, resulting in a simpler implementation;
hence we use this approach in our experimental evaluation.
Suppose one has access to a tuple $(s, \atilde, \rtilde, \stilde')$ 
sampled from a replay buffer with knowledge of the sampling probability
$q(\atilde \mid s)$ (possibly unnormalized) with full support.
Then we draw a {\em phantom} action $a\sim N(\atilde, \var(s))$ 
and optimize $\qtilde_\qparam(s, a)$ by minimizing a weighted Bellman error
\begin{equation}
\frac{1}{q(\atilde|s)} (\qtilde_\qparam(s, a) - \rtilde - \gamma \qtilde_\qparam(\stilde', \mean(\stilde'))^2
~.
\end{equation}
In this way, for a specific pair of state and action $(s, a)$ the expected
objective value is,
\begin{multline}
\hskip-1mm
\expected_{q(\atilde \mid s),\rtilde,\stilde'}
\!
\left[
\delta \!\cdot\!
(\qtilde_\qparam(s, a) 
- \rtilde - \gamma \qtilde_\qparam(\stilde', \mean(\stilde')))^2
\right]
,
\!\!\!
\label{eq:res-qtilde}
\end{multline}
where $\delta = \frac{N(a \mid \atilde,\var(s))}{q(\atilde \mid s)}$.
Note that the denominator of $\delta$ counter-acts the expectation over
$\atilde$ in~\eqref{eq:res-qtilde} and that
the numerator of $\delta$ is 
$N(a|\atilde,\var(s)) = N(\atilde|a,\var(s))$.
Therefore, when the target value $\rtilde + \gamma \qtilde_\qparam(\stilde',\mean(\stilde'))$ 
is treated as fixed (\ie~using a target network) 
this training procedure reaches an optimum when
$\qtilde_\qparam(s,a)$ takes on the target value 
provided in the Bellman equation~\eqref{eq:qtilde2}.

In practice, we find that it is unnecessary to keep track of
the probabilities $q(\atilde\mid s)$, and assume the replay buffer
provides a near-uniform distribution of actions conditioned on states.
Other recent work has also benefited from ignoring or heavily damping
importance weights~\citep{retrace,acer,ppo}.  However, it is
possible when interacting with the environment to save the probability
of sampled actions along with their transitions, and thus have access
to $q(\atilde \mid s)\approx N(\atilde \mid
\mean_{\text{old}}(s),\var_{\text{old}}(s))$.

\subsection{Proximal Policy Optimization}

Policy gradient algorithms are notoriously unstable,
particularly in continuous control problems.
Such instability has motivated the development of trust region methods
that constrain each gradient step to lie within a trust region~\citep{trpo},
or augment the expected reward objective with a penalty on KL-divergence from a 
previous policy~\citep{tpcl,ppo}.
These stabilizing techniques have thus far not been applicable to 
algorithms like DDPG, since the policy is deterministic.
The formulation we propose above,
however, is easily amenable to trust region optimization.
Specifically, we may augment the objective~\eqref{eq:objtilde}
with a penalty
\begin{equation}
\objtrust(\pi) = \objtilde(\pi) - \lambda \int_\S \kl{\pi}{\pi_{\text{old}}} \deriv \rho^\pi(s),
\end{equation}
where $\pi_\text{old}\equiv (\mean_\text{old},\var_\text{old})$
is a previous version of the policy.
The optimization is
straightforward, since the KL-divergence of two Gaussians can be
expressed analytically.

This concludes the technical presentation of the proposed
algorithm {\em Smoothie}:
pseudocode for the full training procedure,
including policy improvement, policy evaluation,
and proximal policy improvement is given in  Algorithm~\ref{alg:ouralg}.
The reader may also refer to the appendix for additional implementation details.

\begin{algorithm}[t] 
\caption{\ouralg}
\label{alg:ouralg}

\begin{algorithmic}
\STATE {\bfseries Input:} 
Environment $ENV$, 
learning rates $\eta_\pi,\eta_Q$, discount factor $\gamma$,
KL-penalty $\lambda$,
batch size $B$, number of training steps $N$,
target network lag $\tau$.

\STATE
\STATE Initialize $\meanparam, \varparam, \qparam$, set $\meanparam'=\meanparam, \varparam'=\varparam, \qparam'=\qparam$.
\FOR{$i=0$ {\bfseries to} $N-1$}
\STATE \emph{// Collect experience}
\STATE Sample action $a\sim N(\mean_{\meanparam}(s), \var_{\varparam}(s))$ and apply to $ENV$ to yield $r$ and $s^\prime$.
\STATE Insert transition $(s, a, r, s^\prime)$ to replay buffer.

\STATE
\STATE \emph{// Train $\mean,\var$}
\STATE Sample batch $\{(s_k,a_k,r_k,s^\prime_k)\}_{k=1}^B$ from replay buffer.
\STATE Compute gradients $g_k = \frac{\partial \qtilde_\qparam(s_k, a)}{\partial a}\at[\big]{a=\mean_{\meanparam}(s_k)}$.
\STATE Compute Hessians $H_k = \frac{\partial^2 \qtilde_\qparam(s_k, a)}{\partial a^2}\at[\big]{a=\mean_{\meanparam}(s_k)}$.
\STATE Compute penalties $KL_k = KL(\mean_{\meanparam},\var_{\varparam} || \mean_{\meanparam'},\var_{\varparam'})$.
\STATE Compute updates 
\STATE $~~~~ \Delta \meanparam = \frac{1}{B}\sum_{k=1}^B g_k \nabla_{\meanparam} \mean_{\meanparam}(s_k) - \lambda \nabla_{\meanparam} KL_k$,
\STATE $~~~~ \Delta \varparam = \frac{1}{B}\sum_{k=1}^B \frac{1}{2}H_k \nabla_{\varparam} \var_{\varparam}(s_k) - \lambda \nabla_{\varparam} KL_k$.
\STATE Update 
$\meanparam \leftarrow \meanparam + \eta_\pi \Delta\meanparam$,
$\varparam \leftarrow \varparam + \eta_\pi \Delta\varparam$.

\STATE
\STATE \emph{// Train $\qtilde$}
\STATE Sample batch $\{(s_k,\atilde_k,\rtilde_k,\stilde^\prime_k)\}_{k=1}^B$ from replay buffer.
\STATE Sample phantom actions $a_k\sim N(\atilde_k, \var_{\varparam}(s_k))$.
\STATE Compute loss 
\STATE $\calL(\qparam) = \frac{1}{B}\sum_{k=1}^B (\qtilde_{\qparam}(s, a) - r - \gamma \qtilde_{\qparam'}(\stilde', \mean_{\meanparam'}(\stilde')))^2$.
\STATE Update $\qparam \leftarrow \qparam - \eta_Q \nabla_\qparam \calL(\qparam)$.

\STATE
\STATE \emph{// Update target variables}
\STATE Update 
$\meanparam'\leftarrow (1 - \tau)\meanparam' + \tau \meanparam$;
$\varparam'\leftarrow (1 - \tau)\varparam' + \tau \varparam$;
$\qparam'\leftarrow (1 - \tau)\qparam' + \tau \qparam$.

\ENDFOR

\end{algorithmic}
\end{algorithm}

\subsection{Compatible Function Approximation}

The approximator $\qtilde_\qparam$ for $\qtilde$ should be sufficiently
accurate so that updates for $\mean_{\meanparam}$ and $\var_{\varparam}$ 
are not affected by substituting 
$\frac{\partial \qtilde_\qparam(s,a)}{\partial a}$ and
$\frac{\partial^2 \qtilde_\qparam(s, a)}{\partial a^2}$ for
$\frac{\partial \qtilde(s,a)}{\partial a}$ and
$\frac{\partial^2 \qtilde(s, a)}{\partial a^2}$
respectively.
Define $\epsilon_k(s, \theta, w)$ as the difference between the 
true $k$-th derivative of $\qtilde$ and the $k$-th derivative
of the approximated $\qtilde_\qparam$ at $a=\mean_{\meanparam}(s)$:
\begin{equation}
\epsilon_k(s, \theta, w) = 
\nabla^k_a\qtilde_\qparam(s, a)\at[\big]{a=\mean_{\meanparam}(s)} - 
\nabla^k_a\qtilde(s, a)\at[\big]{a=\mean_{\meanparam}(s)}.
\end{equation}
We claim that a $\qtilde_\qparam$ is compatible with respect
to $\mean_{\meanparam}$ if
\begin{enumerate}
\item
$\nabla_a \qtilde_\qparam(s, a)\at[\big]{a=\mean_{\meanparam}(s)} = \nabla_{\meanparam}\mean_{\meanparam}(s)^T \qparam$, 
\item
$\nabla_\qparam \int_\S ||\epsilon_1(s, \theta, w)||^2 \deriv \rho^\pi(s) = 0$ 
(\ie\ $\qparam$ minimizes the expected squared error of the gradients).
\end{enumerate}
Additionally, $\qtilde_\qparam$ is compatible with respect
to $\var_{\varparam}$ if
\begin{enumerate}
\item
$\nabla_a^2\qtilde_\qparam(s, a)\at[\big]{a=\mean_{\meanparam}(s)} = \nabla_{\varparam}\var_{\varparam}(s)^T \qparam$, 
\item
$\nabla_\qparam \int_\S ||\epsilon_2(s, \theta, w)||^2 \deriv\rho^\pi(s) = 0$ 
(\ie\ $\qparam$ minimizes the expected squared error of the Hessians).
\end{enumerate}
We provide a proof of these claims in the Appendix.
One possible parameterization of $\qtilde_\qparam$ may be achieved by taking $\qparam = [\qparam_0, \qparam_1, \qparam_2]$
and parameterizing
\begin{align}
\hskip-1mm
\qtilde_\qparam(s, a) 
&= 
V_{\qparam_0}(s) 
+ (a - \mean_{\meanparam}(s))^T \nabla_{\meanparam}\mean_{\meanparam}(s)^T \qparam_1 
\nonumber
\\
&+ (a - \mean_{\meanparam}(s))^T \nabla_{\varparam}\var_{\varparam}(s)^T \qparam_2 (a - \mean_{\meanparam}(s)).
\end{align}

Similar conditions and parameterizations exist for DDPG~\citep{ddpg1},
in terms of $\qexp$.
While it is reassuring to know that there exists a class of function approximators
which are compatible, this fact has largely been ignored 
in practice.
At first glance, it seems 
impossible to satisfy the second set of conditions 
without access to derivative information of the true $\qtilde$
(for DDPG, $\qexp$).
Indeed, the methods for training Q-value approximators 
(equation~\eqref{eq:dq-bellman} and Section~\ref{sec:train-q})
only train to minimize an error between raw scalar values.
For DDPG, we are unaware of any method that allows one to 
train $\qexp_\qparam$ to minimize an error with respect to
the derivatives of the true $\qexp$.
However, the case is different for the smoothed Q-values $\qtilde$.
In fact, it is possible to train $\qtilde_\qparam$
to minimize an error with respect 
to the derivatives of the true $\qtilde$.  
We provide an elaboration in the Appendix.  
In brief, it is possible to use~\eqref{eq:qtilde2} to derive
Bellman-like equations which relate a derivative 
$\frac{\partial^k \qtilde(s, a)}{\partial a^k}$
of any degree $k$ to an integral over the 
raw Q-values at the next time step $\qtilde(\stilde',\mean(\stilde'))$.
Thus, it is possible to devise a training scheme in the spirit of
the one outlined in Section~\ref{sec:train-q}, 
which optimizes 
$\qtilde_\qparam$ to minimize the squared error with the derivatives
of the true $\qtilde$.
This theoretical property of the smoothed Q-values is unique and
provides added benefits to its use over the standard Q-values.

\section{Related Work}

This paper follows a long line of work
that uses Q-value functions to stably learn a policy,
which in the past has been used to
either approximate expected
\citep{sarsa1, sarsa2, gu2017interpolated}
or optimal
\citep{qlearning,ddpg1,pcl,haarnoja2017reinforcement,lmetz}
future value.

Work that is most similar to what we present are methods that
exploit gradient information from the Q-value function to
train a policy.  Deterministic policy gradient~\citep{ddpg1}
is perhaps the best known of these.
The method we propose can be interpreted as a generalization of 
the deterministic policy gradient.  Indeed,
if one takes the limit of the
policy covariance $\Sigma(s)$ as it goes to 0,
the proposed Q-value function becomes the 
deterministic value function of DDPG,
and the updates for training the Q-value approximator and the 
policy mean are identical.

Stochastic Value Gradient (SVG)~\citep{heess2015learning} also 
trains stochastic policies using
an update that is similar to DDPG (\ie\ SVG(0) with replay). 
The key differences with our approach are that SVG does not provide an update
for the covariance, and the mean update in SVG estimates the
gradient with a noisy Monte Carlo sample, 
which we avoid by estimating the smoothed Q-value function.
Although a covariance update could be derived using the same reparameterization
trick as in the mean update, that would also require a noisy Monte Carlo 
estimate.
Methods for updating the covariance 
along the gradient of expected reward are essential for 
applying the subsequent trust region and proximal policy techniques.

More recently, \citet{epg} introduced expected policy gradients (EPG), 
a generalization
of DDPG that provides updates for the mean and covariance of a stochastic
Gaussian policy using gradients of an estimated Q-value function.
In that work, the expected Q-value used in standard
policy gradient algorithms such as SARSA~\citep{suttonbook, sarsa1, sarsa2}
is estimated.
The updates in EPG therefore require approximating an integral of the expected
Q-value function, or assuming the Q-value has a simple form that allows for
analytic computation. 
Our analogous process directly estimates an integral
(via the smoothed Q-value function) and avoids
approximate integrals, thereby making the updates simpler and generally applicable. 
Moreover, while~\citet{epg} 
rely on a quadratic Taylor expansion of the estimated Q-value function, 
we instead
rely on the strength of neural network function approximators 
to directly estimate
the smoothed Q-value function.


The novel training scheme we propose for learning 
the covariance of a Gaussian policy relies on properties of 
Gaussian integrals~\citep{bonnet1964transformations,price1958useful}. 
Similar identities have been used in the past to derive updates for 
variational auto-encoders~\citep{kingma2013auto}
and Gaussian back-propagation~\citep{rezende2014stochastic}.

Finally, the perspective presented in this paper, where Q-values represent
the averaged return of a distribution of actions rather than
a single action, is distinct from recent 
advances in distributional RL~\citep{distributional}.
Those approaches focus on the distribution of returns of a single action,
whereas we consider the single average return of a distribution of actions.  
Although we restrict our attention in this paper to Gaussian policies,
an interesting topic for further investigation is to
study the applicability of this new perspective to a
wider class of policy distributions.

\section{Experiments}

\begin{figure}[t]
\begin{center}
  \begin{tabular}{c}
\hskip-3mm
\includegraphics[width=1.03\columnwidth]{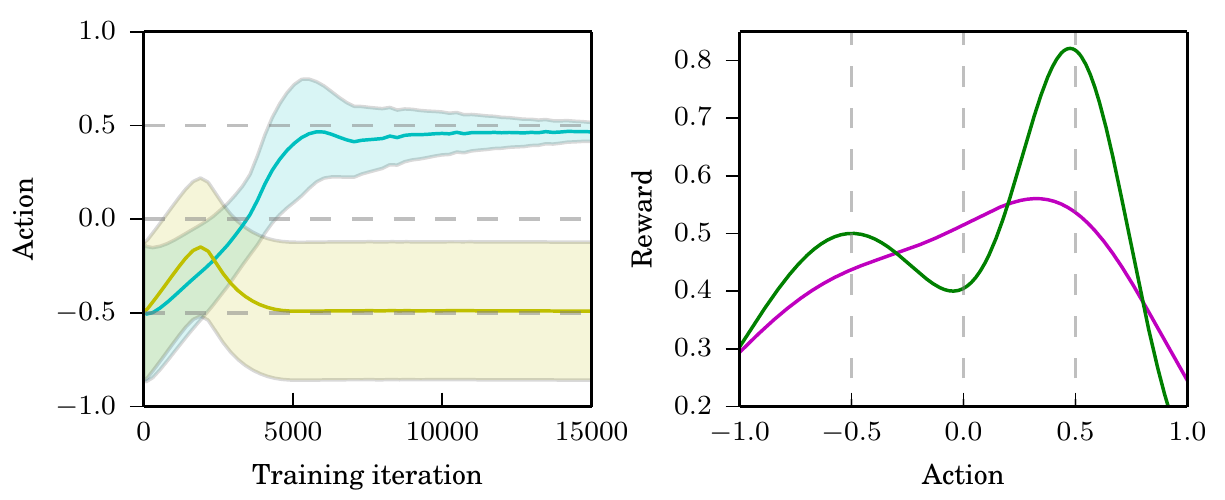} \\
    \multicolumn{1}{c}{~\includegraphics[width=0.8\columnwidth]{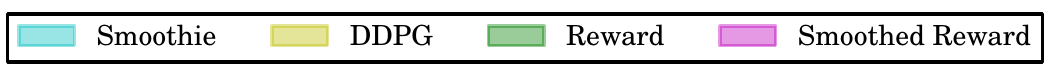}}
  \end{tabular}
\end{center}
\caption{
Left:
The learnable policy mean and standard deviation during training 
for \ouralg and DDPG on the simple synthetic task introduced in 
\secref{sec:idea}.  
The standard deviation for DDPG is the exploratory noise kept constant
during training.
Right:
Copy of \figref{fig:bumps} showing the reward function
and its Gaussian-smoothed version.
\ouralg successfully escapes the lower-reward local optimum,
while increasing then decreasing its policy variance as the convexity/concavity
of the smoothed reward function changes.
}
\vspace{-0.3in}
\label{fig:toy}
\end{figure}

We utilize the insights 
from Section~\ref{sec:method}
to introduce a new RL algorithm, {\em \ouralg}.
\ouralg maintains a parameterized $\qtilde_\qparam$
trained via the procedure described in Section~\ref{sec:train-q}.
It then uses the gradient and Hessian of this approximation
to train a Gaussian policy
$\pi_{\meanparam,\varparam}\equiv(\mean_{\meanparam},\var_{\varparam})$
using the updates stated in~\eqref{eq:mean-upd} 
and~\eqref{eq:var-upd}.
See Algorithm~\ref{alg:ouralg} for a simplified pseudocode of the algorithm.

We perform a number of evaluations of \ouralg to compare to DDPG.  
We choose DDPG as a baseline because it 
(1) utilizes gradient information of a Q-value approximator,
much like the proposed algorithm; 
and (2) is a standard algorithm well-known to have
achieve good, sample-efficient performance on 
continuous control benchmarks.

\subsection{Synthetic Task}
\label{sec:toy}

Before investigating benchmark problems,
we first briefly revisit the simple 
task introduced
in Section~\ref{sec:idea} and reproduced in~\figref{fig:toy} (Right).
Here, the reward function is a mixture of two Gaussians,
one better than the other, and  we initialize the policy mean to be
centered on the worse of the two. 
We plot the learnable policy mean and standard deviation during training for
\ouralg and DDPG in~\figref{fig:toy} (Left).
\ouralg learns both the mean and variance, while DDPG
learns only the mean and the variance plotted is the 
exploratory noise, whose scale is kept fixed during training.

\begin{figure}[t!]
\begin{center}
  \begin{tabular}{@{}c@{\hspace*{.4cm}}c@{}}
    \tiny HalfCheetah & \tiny Swimmer \\
    \includegraphics[width=0.4\columnwidth]{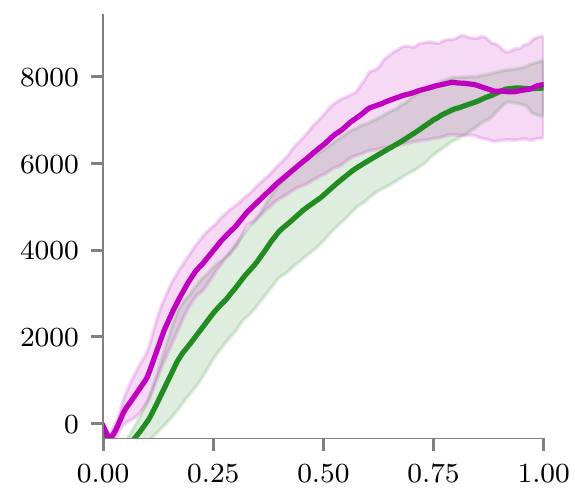} &
    \includegraphics[width=0.4\columnwidth]{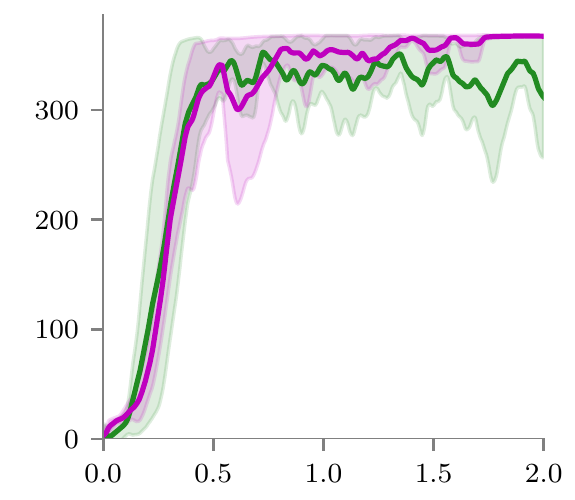} \\
    \tiny Hopper & \tiny Walker2d \\
    \includegraphics[width=0.4\columnwidth]{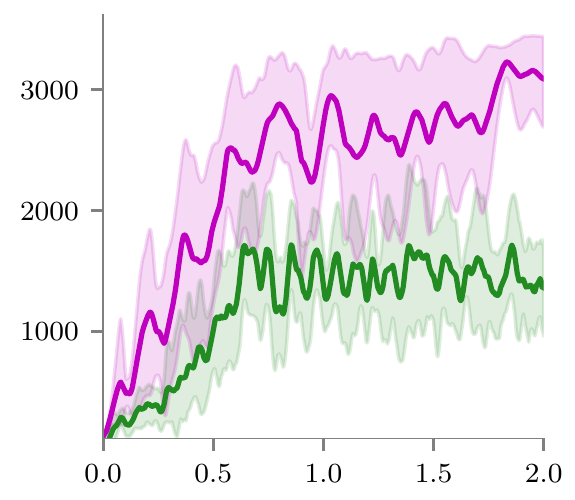} &
    \includegraphics[width=0.4\columnwidth]{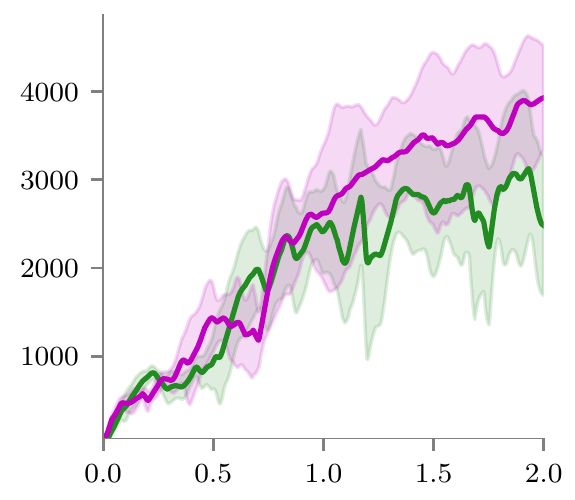} \\
    \tiny Ant & \tiny Humanoid \\
    \includegraphics[width=0.4\columnwidth]{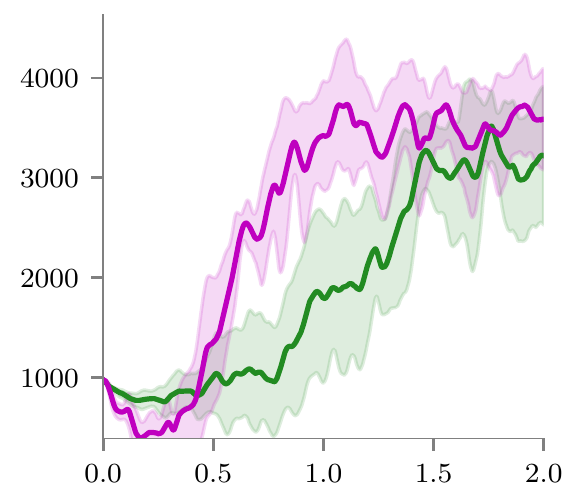} &
    \includegraphics[width=0.4\columnwidth]{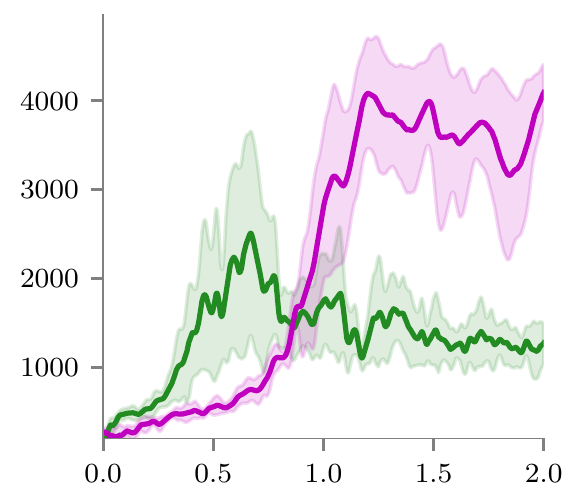} \\
    \multicolumn{2}{c}{\includegraphics[width=0.8\columnwidth]{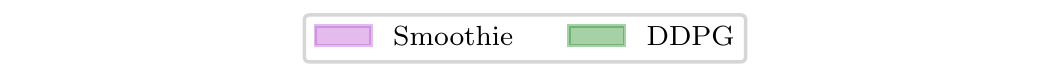}}
  \end{tabular}
\end{center}
\caption{
Results of \ouralg and DDPG on 
continuous control benchmarks. The x-axis is
in millions of environment steps.
Each plot shows the average reward and standard deviation clipped at the min and max
of six randomly seeded runs after choosing best hyperparameters.
We see that \ouralg is competitive with DDPG even when DDPG uses a
hyperparameter-tuned noise scale, 
and
\ouralg learns the optimal
noise scale (the covariance) during training.
Moreoever, we observe significant advantages in terms of final
reward performance, especially in the more
difficult tasks like Hopper, Walker2d, and Humanoid.
}
\label{fig:results}
\end{figure}

As expected, we observe that DDPG cannot escape the local optimum.
At the beginning of training it exhibits some movement away from 
the local optimum
(likely due to the initial noisy approximation given by $\qdet_\qparam$).
However, it is unable to progress very far from the initial mean.
Note that this is not an issue of exploration.  The exploration scale
is high enough that $\qdet_\qparam$ is aware of the better Gaussian.
The issue is in the update for $\mean_{\meanparam}$, 
which is only with regard to
the derivative of $\qdet_\qparam$ at the current mean.

On the other hand, we find \ouralg is easily able to solve the task.
This is because the smoothed reward function approximated by 
$\qtilde_\qparam$ has a derivative that clearly points $\mean_{\meanparam}$
toward the better Gaussian.
We also observe that \ouralg is able to suitably adjust the covariance
$\var_{\varparam}$ during training.
Initially, $\var_{\varparam}$ decreases due to the concavity
of the smoothed reward function.
As a region of convexity is entered, it begins to increase,
before again decreasing to near-zero as $\mean_{\meanparam}$
approaches the global optimum.
This example clearly shows the ability of \ouralg to exploit the smoothed
action value landscape.

\subsection{Continuous Control}

Next, we
consider
standard continuous control benchmarks
available on OpenAI Gym~\citep{gym}
utilizing the MuJoCo environment~\citep{mujoco}.

Our implementations utilize feed forward neural networks for policy
and Q-values.  We parameterize the covariance $\var_{\varparam}$
as a diagonal given by $e^{\varparam}$.  The exploration for DDPG
is determined by an
Ornstein-Uhlenbeck process~\citep{uhlenbeck,ddpg2}.
Additional implementation details are provided in the Appendix.

\comment{
\begin{figure}[h]
\begin{center}
  \begin{tabular}{@{}c@{\hspace*{.4cm}}c@{\hspace*{.4cm}}c@{}}
    \tiny HalfCheetah & \tiny Swimmer & \tiny Hopper \\
    \includegraphics[width=0.27\columnwidth]{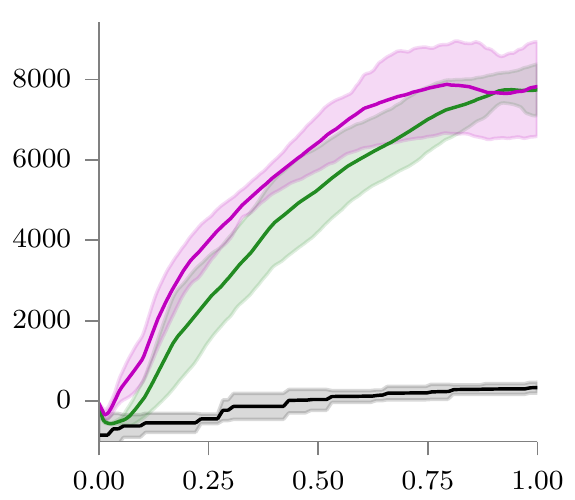} &
    \includegraphics[width=0.27\columnwidth]{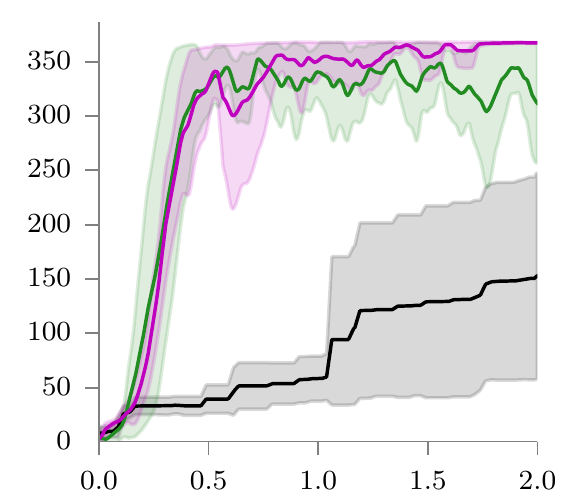} &
    \includegraphics[width=0.27\columnwidth]{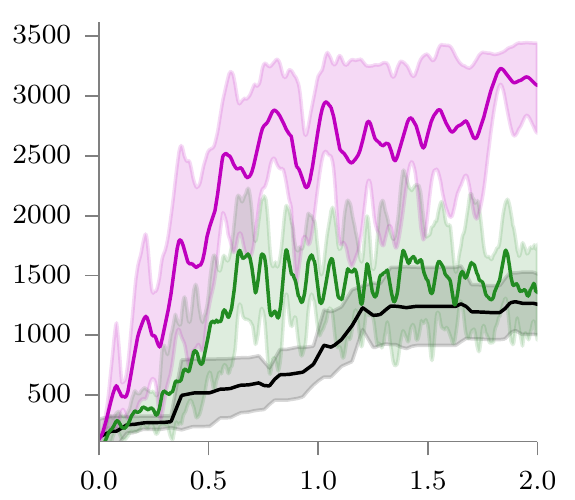} \\
    \tiny Walker2d & \tiny Ant & \tiny Humanoid \\
    \includegraphics[width=0.27\columnwidth]{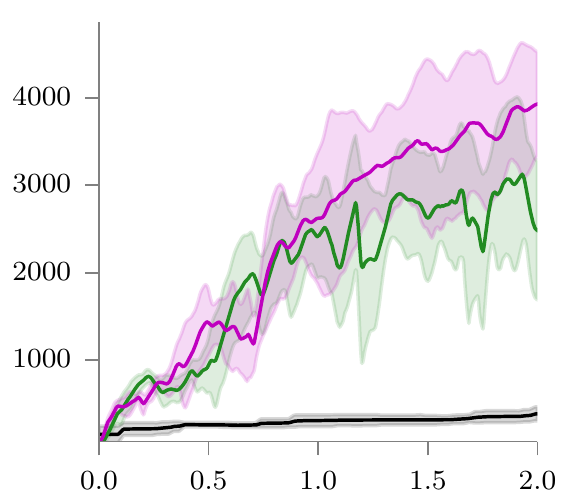} &
    \includegraphics[width=0.27\columnwidth]{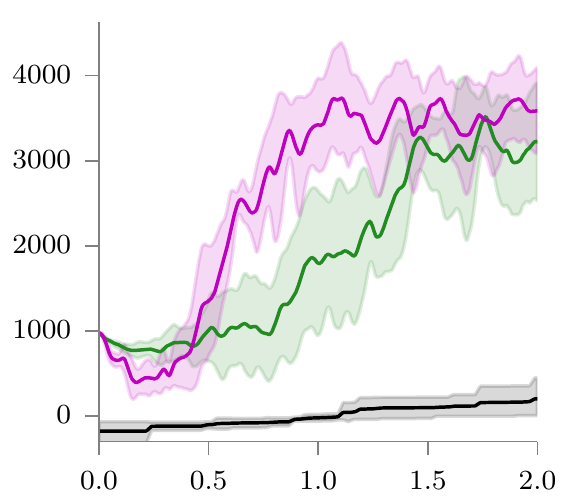} &
    \includegraphics[width=0.27\columnwidth]{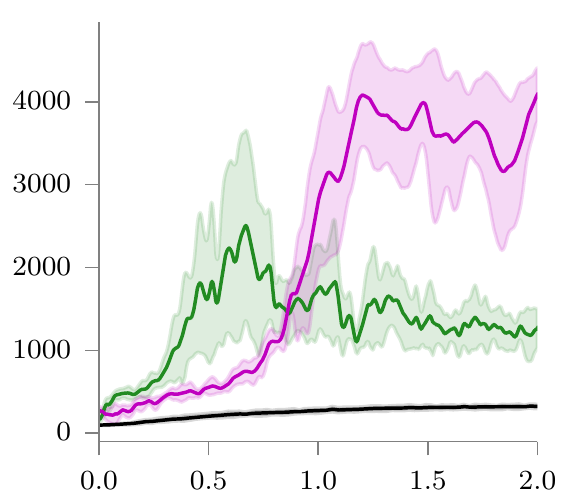} \\
    \multicolumn{3}{c}{\includegraphics[width=0.5\columnwidth]{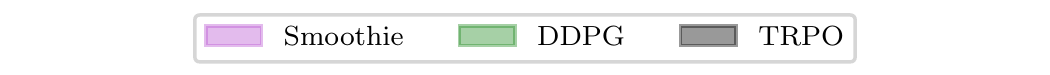}}
  \end{tabular}
\end{center}
\caption{
Results of \ouralg, DDPG, and TRPO on 
continuous control benchmarks. The x-axis is
in millions of environment steps.
Each plot shows the average reward and standard deviation clipped at the min and max
of six randomly seeded runs after choosing best hyperparameters.
We see that \ouralg is competitive with DDPG even when DDPG uses a
hyperparameter-tuned noise scale, 
and
\ouralg learns the optimal
noise scale (the covariance) during training.
Moreoever, we observe significant advantages in terms of final
reward performance, especially in the more
difficult tasks like Hopper, Walker2d, and Humanoid.
Across all tasks, TRPO is not sufficiently sample-efficient
to provide a competitive baseline.
}
\label{fig:results}
\end{figure}
}

We compare the results of \ouralg and DDPG in~\figref{fig:results}.
For each task we performed a hyperparameter search over 
actor learning rate, critic learning rate and reward scale, and 
plot the average of six runs for the best hyperparameters.
For DDPG we extended the hyperparameter search to also consider
the scale and damping of exploratory noise provided by the
Ornstein-Uhlenbeck process.  \ouralg, on the other hand,
contains an additional hyperparameter to determine
the weight on KL-penalty.

Despite DDPG having the advantage of its exploration
decided by a hyperparameter search while \ouralg must
learn its exploration without supervision, we find
that \ouralg performs competitively or better across all tasks,
exhibiting a slight advantage in Swimmer and Ant,
while showing more dramatic improvements in Hopper, Walker2d, and Humanoid.
The improvement is especially dramatic
for Hopper, where the average reward is doubled.
We also highlight the results for Humanoid, which
as far as we know, are the
best published results for a method that only trains
on the order of millions of environment steps.
In contrast, TRPO, which to the best of our knowledge is the
only other algorithm that can achieve competitive 
performance,
requires on the order of tens of millions of environment steps to achieve
comparable reward.
This gives added evidence to the benefits of using 
a learnable covariance and not restricting a policy to 
be deterministic.

\begin{figure}[t]
\begin{center}
  \begin{tabular}{@{}c@{\hspace*{.4cm}}c@{}}
    \tiny Hopper & \tiny Walker2d \\
    \includegraphics[width=0.4\columnwidth]{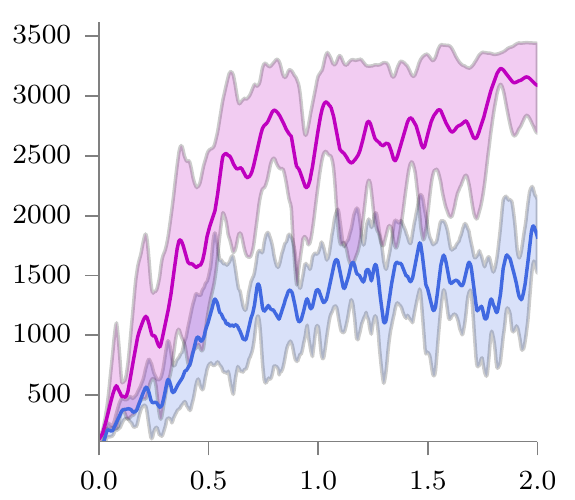} &
    \includegraphics[width=0.4\columnwidth]{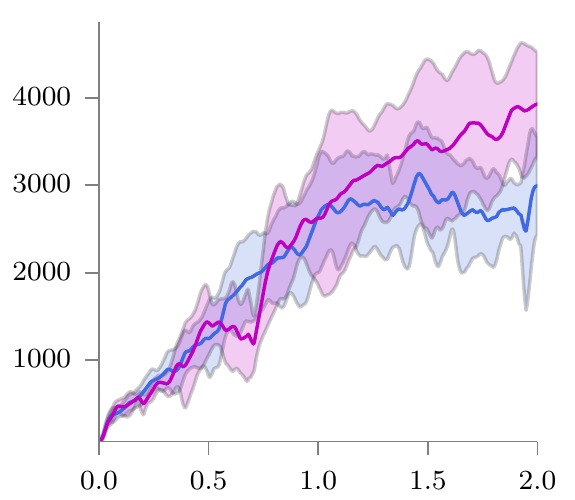} \\
    \tiny Ant & \tiny Humanoid \\
    \includegraphics[width=0.4\columnwidth]{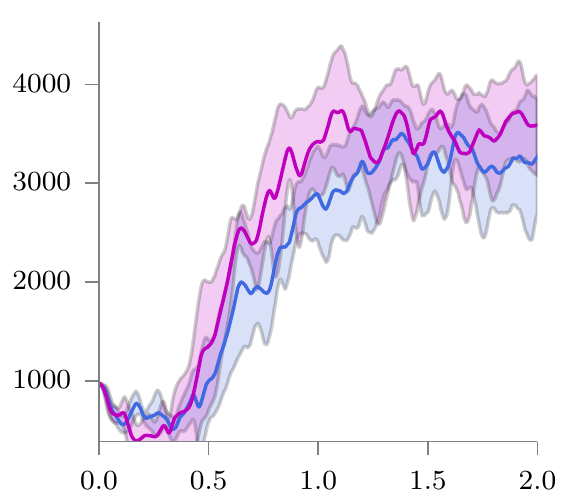} &
    \includegraphics[width=0.4\columnwidth]{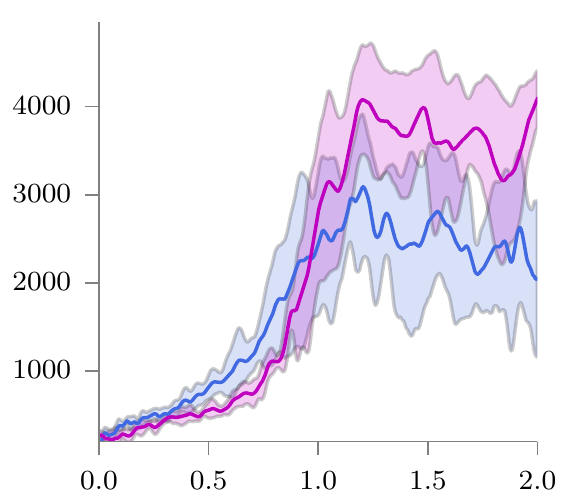} \\
    \multicolumn{2}{c}{\includegraphics[width=0.8\columnwidth]{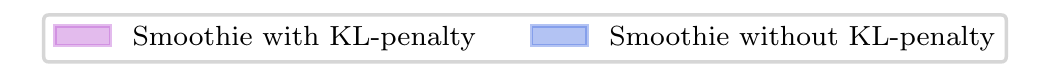}}
  \end{tabular}
\end{center}
\caption{
Results of \ouralg with and without a KL-penalty. The x-axis is
in millions of environment steps.  
We observe benefits of using a proximal policy optimization method, especially
in Hopper and Humanoid, where the performance improvement is significant
without sacrificing sample efficiency.
}
\vspace{-0.2in}
\label{fig:trust}
\end{figure}

\comment{
These results make it clear that on more difficult continuous
control tasks (Hopper, Walker2d, Ant, Humanoid), 
both DDPG and \ouralg suffer from instability,
showing high variance across runs and a performance that is 
far behind trust region and proximal policy methods~\citep{trpo, tpcl, ppo}.
Thus, we evaluated \ouralg with a KL-penalty and present 
these results in~\figref{fig:trust}.
}

Empirically, we found the introduction of a KL-penalty to improve performance
of \ouralg, especially on harder tasks.  
We present a comparison of results of \ouralg with and without
the KL-penalty on the four harder tasks in~\figref{fig:trust}.
A KL-penalty to encourage stability is not possible in DDPG.
Thus, \ouralg provides a much needed solution to
the inherent instability in DDPG training.

\comment{
We observe that \ouralg augmented with a KL-penalty consistently
improves performance.  The improvement is especially dramatic
for Hopper, where the average reward is doubled.
We also highlight the results for Humanoid, which
also exhibits a significant improvement.
As far as we know, these Humanoid results are the 
best published results for a method that only trains
on the order of millions of environment steps.
In contrast, TRPO, which to the best of our knowledge is the
only other algorithm which can achieve better performance,
requires on the order of tens of millions of environment steps.
This gives added evidence to the benefits of using 
a learnable covariance and not restricting a policy to 
be deterministic.
}

\section{Conclusion}

We have presented a new Q-value function concept, $\qtilde$, 
that is a Gaussian-smoothed version of the standard expected Q-value, $\qexp$.
The advantage of $\qtilde$ over $\qexp$ is that its gradient and Hessian
possess an intimate relationship with the gradient of expected reward
with respect to mean and covariance of a Gaussian policy.
The resulting algorithm, \ouralg, is able to successfully
learn both mean and covariance during training, leading to
performance that surpasses that of DDPG,
especially when incorporating a penalty on divergence from a previous policy.

\comment{
\citet{gu2017interpolated} showed that DDPG sits on one end of a spectrum of methods that
interpolate between off-policy updates and on-policy policy gradient updates. Future 
work could determine if Smoothie can also benefit from the improved stability
of interpolated methods.
}

The success of $\qtilde$ is encouraging.  
Intuitively it
appears advantageous to learn $\qtilde$ instead of $\qexp$.
The smoothed Q-values by definition
make the true reward surface smoother, thus
possibly easier to learn; moreover
the smoothed Q-values have a more direct 
relationship with the expected discounted return objective.
We encourage 
further 
investigation of
these claims 
and techniques for applying
the underlying motivations
for $\qtilde$ to other types of policies.

\section{Acknowledgments}

We thank Luke Metz, Sergey Levine, and the Google Brain
team for insightful comments and discussions.


\newpage

\bibliography{main.bib}
\bibliographystyle{icml2018}

\onecolumn
\newpage
\appendix

\section{Proof of Theorem~1}
We want to show that for any $s, a$,
\begin{equation}
\frac{\partial \qtilde(s, a)}{\partial \var(s)}
=
\frac{1}{2} \cdot \frac{\partial^2 \qtilde(s, a)}{\partial a^2}
\label{eq:sigma-trick2}
\end{equation}

We note that similar identities for Gaussian integrals exist in the 
literature \citep{price1958useful,rezende2014stochastic} and point the reader
to these works for further information.  

{\bf Proof.}
The specific identity we state 
may be derived using standard matrix calculus.
We make use of the fact that
\begin{equation}
\frac{\partial}{\partial A} |A|^{-1/2} = -\frac{1}{2}|A|^{-3/2} \frac{\partial}{\partial A} |A| = -\frac{1}{2} |A|^{-1/2} A^{-1},
\end{equation}
and for symmetric $A$,
\begin{equation}
\frac{\partial}{\partial A} ||v||_{A^{-1}}^2 = -A^{-1} vv^T A^{-1}.
\end{equation}
We omit $s$ from $\var(s)$ in the following equations for succinctness.
The LHS of~\eqref{eq:sigma-trick2} is
%
%
\begin{eqnarray*}
\lefteqn{
\int_{\A} \qexp(s, \atilde) \frac{\partial}{\partial \var}
N(\atilde|a,\var) \deriv \atilde
}
	\\[1ex]
& = &
\int_{\A} \qexp(s, \atilde)
\exp\left\{-\frac{1}{2}||\atilde - a||_{\var^{-1}}^2\right\}
\left(
\frac{\partial}{\partial \var} |2\pi\var|^{-1/2} -
\frac{1}{2}|2\pi\var|^{-1/2}
\frac{\partial}{\partial \var} ||\atilde - a||_{\var^{-1}}^2
\right) \deriv \atilde
	\\[1ex]
& = &
\frac{1}{2} \int_{\A} \qexp(s, \atilde) N(\atilde | a, \var)
\left(-\var^{-1} + \var^{-1}(\atilde-a)(\atilde-a)^T \var^{-1}\right) \deriv \atilde.
\end{eqnarray*}

Meanwhile, towards tackling the RHS of~\eqref{eq:sigma-trick2} we
note that
\begin{equation}
\frac{\partial\qtilde(s, a)}{\partial a}
=
\int_{\A} \qexp(s, \atilde) N(\atilde|a,\var) \var^{-1}
(\atilde-a) \deriv\atilde
\;
.
\end{equation}
Thus we have
\begin{eqnarray*}
\frac{\partial^2\qtilde(s, a)}{\partial a^2}
& = &
\int_{\A} \qexp(s, \atilde) \left(
\var^{-1} (\atilde-a) \frac{\partial}{\partial a} N(\atilde|a,\var) +
N(\atilde|a,\var) \frac{\partial}{\partial a} \var^{-1} (\atilde-a)
\right)\deriv \atilde
	\\[1ex]
& = &
\int_{\A} \qexp(s, \atilde) N(\atilde|a,\var)
(\var^{-1}(\atilde-a)(\atilde-a)^T \var^{-1} - \var^{-1})\,\deriv\atilde
\;
.
\end{eqnarray*}
%
%
\hfill$\blacksquare$

\section{Compatible Function Approximation}
\comment{
A function approximator $\qtilde_\qparam$ of $\qtilde$ should be sufficiently
accurate so that updates for $\mean_{\meanparam},\var_{\varparam}$ are not affected
by substituting
$\frac{\partial \qtilde_\qparam(s,a)}{\partial a}$ and
$\frac{\partial^2 \qtilde_\qparam(s, a)}{\partial a^2}$ for
$\frac{\partial \qtilde(s,a)}{\partial a}$ and
$\frac{\partial^2 \qtilde(s, a)}{\partial a^2}$,
respectively.
}

We claim that a $\qtilde_\qparam$ is compatible with respect
to $\mean_{\meanparam}$ if
\begin{enumerate}
\item
$\nabla_a \qtilde_\qparam(s, a)\at[\big]{a=\mean_{\meanparam}(s)}
= \nabla_{\meanparam}\mean_{\meanparam}(s)^T \qparam$,
\item
$
\nabla_\qparam
\displaystyle
\int_\S
\Big( \nabla_a\qtilde_\qparam(s, a)\at[\big]{a=\mean_{\meanparam}(s)} -
\nabla_a\qtilde(s, a)\at[\big]{a=\mean_{\meanparam}(s)} \Big)^2
\deriv \rho^\pi(s)
= 0
$
\quad
(\ie\ $\qparam$ minimizes the expected squared error of the gradients).
\end{enumerate}
Additionally, $\qtilde_\qparam$ is compatible with respect
to $\var_{\varparam}$ if
\begin{enumerate}
\item
$\nabla_a^2\qtilde_\qparam(s, a)\at[\big]{a=\mean_{\meanparam}(s)} = \nabla_{\varparam}\var_{\varparam}(s)^T \qparam$,
\item
$
\nabla_\qparam
\displaystyle
\int_\S
\Big( \nabla_a^2 \qtilde_\qparam(s, a)\at[\big]{a=\mean_{\meanparam}(s)} -
\nabla_a^2 \qtilde(s, a)\at[\big]{a=\mean_{\meanparam}(s)} \Big)^2
\deriv\rho^\pi(s)
= 0
$
\quad
(\ie\ $\qparam$ minimizes the expected squared error of the Hessians).
\end{enumerate}

{\bf Proof.}
We shall show how the conditions stated for compatibility with respect to
$\var_{\varparam}$ are sufficient.
The reasoning for $\mu_{\meanparam}$ follows via a similar argument.
We also refer the reader to~\citet{ddpg1}
which includes a similar procedure for showing compatibility.

From the second condition for compatibility with respect to
$\var_{\varparam}$ we have
\begin{eqnarray*}
\int_\S
\left( \nabla_a^2 \qtilde_\qparam(s, a)\at[\big]{a=\mean_{\meanparam}(s)} - \nabla_a^2 \qtilde(s,a)\at[\big]{a=\mean_{\meanparam}(s)} \right) 
\nabla_\qparam\left(\nabla_a^2 \qtilde_\qparam(s, a)\at[\big]{a=\mean_{\meanparam}(s)} \right)
\;
\deriv\rho^\pi(s)
\;\;\; = \;\;\; 0
\;
.
\end{eqnarray*}
We may combine this with the first condition to find
\begin{eqnarray*}
\int_\S \nabla_a^2 \qtilde_\qparam(s, a)\at[\big]{a=\mean_{\meanparam}(s)} \nabla_{\varparam}\var_{\varparam}(s) \deriv\rho^\pi(s)
	& = &
\int_\S \nabla_a^2 \qtilde(s, a)\at[\big]{a=\mean_{\meanparam}(s)} \nabla_{\varparam}\var_{\varparam}(s) \deriv\rho^\pi(s)
\;
,
\end{eqnarray*}
which is the desired property for compatibility.\hfill$\blacksquare$

\section{Derivative Bellman Equations}
The conditions for compatibility require training $\qtilde_\qparam$ to fit the true $\qtilde$
with respect to derivatives.  Howevever, in RL contexts, one often does not have access
to the derivatives of the true $\qtilde$.  In this section, we elaborate on a method
to train $\qtilde_\qparam$ to fit the derivatives of the true $\qtilde$ without
access to true derivative information.

Our method relies on a novel formulation: {\em derivative Bellman equations}.
We begin with the standard $\qtilde$ Bellman equation presented in the main paper:
\begin{equation}
\qtilde(s, a) = \int_{\A}
N(\atilde \mid a,\var(s))\,\expected_{\rtilde,\stilde'}
\left[\rtilde + \gamma \qtilde(\stilde', \mean(\stilde')) \right]
\deriv \atilde~.
\end{equation}
One may take derivatives of both sides to yield the following identity for any $k$:
\begin{equation}
\frac{\partial^k \qtilde(s, a)}{\partial a^k} = \int_{\A}
\frac{\partial^k N(\atilde \mid a,\var(s))}{\partial a^k}\,\expected_{\rtilde,\stilde'}
\left[\rtilde + \gamma \qtilde(\stilde', \mean(\stilde')) \right]
\deriv \atilde~.
\end{equation}
One may express the $k$-the derivative of a normal density for $k\le 2$ simply as
\begin{equation}
\frac{\partial^k N(\atilde \mid a,\var(s))}{\partial a^k} =
N(\atilde \mid a,\var(s)) \var(s)^{-k/2} \cdot H_k(\var(s)^{-1/2}(\atilde - a)),
\end{equation}
where $H_k$ is a polynomial.  Therefore, we have the following derivative
Bellman equations for any $k\le 2$:
\begin{equation}
\frac{\partial^k \qtilde(s, a)}{\partial a^k} = \int_{\A}
N(\atilde \mid a,\var(s)) \var(s)^{-k/2} \cdot H_k(\var(s)^{-1/2}(\atilde - a)) \,
\expected_{\rtilde,\stilde'}
\left[\rtilde + \gamma \qtilde(\stilde', \mean(\stilde')) \right]
\deriv \atilde~.
\label{eq:deriv-bellman}
\end{equation}

One may train a parameterized $\qtilde_\qparam$ to satisfy these consistencies
in a manner similar to that described in Section 4.2. 
Specifically, suppose one has access to a tuple $(s, \atilde, \rtilde, \stilde')$
sampled from a replay buffer
with knowledge of the sampling probability $q(\atilde \mid s)$ (possibly unnormalized)
with full support.
Then
we draw a {\em phantom} action $a\sim N(\atilde, \var(s))$
and optimize $\qtilde_\qparam(s, a)$
by minimizing a weighted derivative Bellman error
\begin{equation}
\frac{1}{q(\atilde|s)} \left(\frac{\partial^k \qtilde_\qparam(s, a)}{\partial a^k}
- \var(s)^{-k/2} \cdot H_k(\var(s)^{-1/2}(a - \atilde)) (\rtilde + \gamma \qtilde_\qparam(\stilde', \mean(\stilde')))\right)^2,
\end{equation}
for $k=0, 1, 2$.
As in the main text, it is possible to argue that
when
using target networks,
this training procedure reaches an optimum when
$\qtilde_\qparam(s,a)$ satisfies
the recursion in the derivative Bellman equations~\eqref{eq:deriv-bellman}
for $k=0, 1, 2$.

\comment{
\section{Additional Plots}
We present an additional comparison with TRPO in~\figref{fig:trpo}.
\begin{figure}[h]
\begin{center}
  \begin{tabular}{@{}c@{}c@{}c@{}c@{}}
    \tiny Hopper & \tiny Walker2d & \tiny Ant & \tiny Humanoid \\
    \includegraphics[width=0.23\columnwidth]{figs/trpo_rewards2} &
    \includegraphics[width=0.23\columnwidth]{figs/trpo_rewards3} &
    \includegraphics[width=0.23\columnwidth]{figs/trpo_rewards4} &
    \includegraphics[width=0.23\columnwidth]{figs/trpo_rewards5} \\
    \multicolumn{4}{c}{\includegraphics[width=0.5\columnwidth]{figs/trpo_rewards-1}}
  \end{tabular}
\end{center}
\caption{
Results of \ouralg augmented with a KL-penalty, DDPG, and TRPO 
with x-axis in millions of environment steps.
}
\label{fig:trpo}
\end{figure}
}

\section{Implementation Details}

We utilize feed forward networks for both policy and Q-value approximator.
For $\mean_{\meanparam}(s)$ we use two hidden layers of dimensions $(400, 300)$
and relu activation functions.
For $\qtilde_{\qparam}(s, a)$ and $\qexp_{\qparam}(s, a)$
we first embed the state into a 400 dimensional vector
using a fully-connected layer and $\tanh$ non-linearity.  We then
concatenate the embedded state with $a$ and pass the result through
a 1-hidden layer neural network of dimension $300$ with $\tanh$ activations.
We use a diagonal $\var_{\varparam}(s) = e^\varparam$ for \ouralg, with $\varparam$
initialized to $-1$.

To find optimal hyperparameters we perform a 100-trial random search
over the hyperparameters specified in~\tabref{tab:hparams}.
The OU exploration parameters only apply to DDPG.
The $\lambda$ coefficient on KL-penalty only applies to \ouralg with a KL-penalty.

\begin{table}[t]
\centering
\begin{tabular}{|c|c|c|}
\hline
{\bf Hyperparameter} & {\bf Range} & {\bf Sampling} \\
\hline 
actor learning rate & [1e-6,1e-3] & log \\
critic learning rate & [1e-6,1e-3] & log \\
reward scale & [0.01,0.3] & log \\
OU damping & [1e-4,1e-3] & log \\
OU stddev & [1e-3,1.0] & log \\
$\lambda$ & [1e-6, 4e-2] & log \\
discount factor & 0.995 & fixed \\
target network lag & 0.01 & fixed \\
batch size & 128 & fixed \\
clipping on gradients of $Q$ & 4.0 & fixed \\
num gradient updates per observation & 1 & fixed \\
Huber loss clipping & 1.0 & fixed \\
\hline
\end{tabular}
\caption{
Random hyperparameter search procedure.
We also include the hyperparameters which we kept fixed.
}
\label{tab:hparams}
\end{table}

\subsection{Fast Computation of Gradients and Hessians}
The Smoothie algorithm relies on the computation of the gradients 
$\frac{\partial \qtilde_\qparam(s, a)}{\partial a}$ and Hessians 
$\frac{\partial^2 \qtilde_\qparam(s, a)}{\partial a^2}$.
In general, these quantities may be computed through multiple backward passes of a 
computation graph.  
However, for faster training, in our implementation we take advantage of a 
more efficient computation.
We make use of the following identities:
\begin{equation}
\frac{\partial}{\partial x} f(g(x)) = f'(g(x)) \frac{\partial}{\partial x} g(x),
\end{equation}
\begin{equation}
\frac{\partial^2}{\partial x^2} f(g(x)) = 
\left(\frac{\partial}{\partial x} g(x)\right)^T f''(g(x)) \frac{\partial}{\partial x} g(x) +
f'(g(x)) \frac{\partial^2}{\partial x^2} g(x).
\end{equation}
Thus, during the forward computation of our critic network $\qtilde_\qparam$, we not only maintain the tensor output $O_L$ of layer $L$, but also the tensor $G_L$ corresponding to the gradients of $O_L$ with respect to input actions and the tensor $H_L$ corresponding to the Hessians of $O_L$ with respect to input actions.
At each layer we may compute $O_{L+1}, G_{L+1}, H_{L+1}$ given $O_L,G_L,H_L$.
Moreover, since we utilize feed-forward fully-connected layers, the computation of
$O_{L+1}, G_{L+1}, H_{L+1}$ may be computed using fast tensor products.

\end{document}